# Fractional differentiation based image processing

**Amelia Carolina Sparavigna**
Dipartimento di Fisica, Politecnico di Torino
C.so Duca degli Abruzzi 24, Torino, Italy

**Abstract**
There are many resources useful for processing images, most of them freely available and quite friendly to use. In spite of this abundance of tools, a study of the processing methods is still worthy of efforts. Here, we want to discuss the new possibilities arising from the use of fractional differential calculus. This calculus evolved in the research field of pure mathematics until 1920, when applied science started to use it. Only recently, fractional calculus was involved in image processing methods. As we shall see, the fractional calculation is able to enhance the quality of images, with interesting possibilities in edge detection and image restoration. We suggest also the fractional differentiation as a tool to reveal faint objects in astronomical images.

**1. Introduction**
The fractional differentiation has started to play a very important role in various research fields, also for image and signal processing. In image processing, the fractional calculus can be rather interesting for filtering and edge detection, giving a new approach to enhance the quality of images. Fractional calculus is generalizing derivative and integration of a function to non-integer order [1-3]. As discussed by many researchers, probably a name as "generalized calculus" would be better than "fractional", which is actually used.
We are in fact familiar with notations $D^1 f(x)$ and $D^2 f(x)$ for first and second order derivatives, but many people can be in doubt about the meaning of the notation $D^{1/2} f(x)$, describing the 1/2- order derivative. People, to which notation looks rather unfamiliar, could gain a misleading idea that this is a calculus just recently developed. In fact, the problem of fractional derivatives is rather old. Leibnitz already discussed it in eighteenth century and other famous names of the past studied and contributed to the development of fractional calculus in the field of pure mathematics [4]. We see the first applications of fractional calculus in 1920 and only recently, it was applied to image processing [5].
This paper discusses how the fractional calculus can provide benefits to image processing. In particular, we will see that it is useful in edge detection and for enhancing the image quality. The paper ends with examples on astronomical images.

**2. Discrete fractional derivative**
Allowing integration and derivation of any positive real order, the fractional calculus can be considered a branch of mathematical analysis, which deals with integro-differential equations. Then, the calculus of derivatives is not straightforward as the calculus of integer order derivatives. It is quite complex but the reader can find concise descriptions of this calculus in Ref.[6] and [7].
Since image processing is usually working on quantized and discrete data, we discuss just the discrete implementation of fractional derivation. We have to deal with two-dimensional image maps, which are two-dimensional arrays of pixels, each pixel having three colour tones. The recorded image is a discrete signal, because it is discrete the position of the pixel in the map. Moreover, the signal is quantized, since the colour tones are ranging from 0 to 255. To define the partial derivatives suitable for calculations on the image maps, let us define the discrete fractional differentiation in the following way.

As in Ref.[8] and [9], let us suppose to have a signal $s(t)$, where $t$ can have only discrete values, $t = 1,2,...,n$: the fractional differentiation of this signal is given by:

$$\frac{d^\nu s(t)}{dt^\nu} = s(t) + (-\nu)s(t-1) + \frac{(-\nu)(-\nu+1)}{2}s(t-2) + \frac{(-\nu)(-\nu+1)(-\nu+2)}{6}s(t-3) + ...$$

(1)

where $\nu$ is a real number. If we have a bidimensional map $s(x,y)$, where $x,y$ can have only discrete values, that is $x = 1,2,...,n_x$ and $y = 1,2,...,n_y$, the partial derivatives are:

$$\frac{\partial^\nu s(x,y)}{\partial x^\nu} = s(x,y) + (-\nu)s(x-1,y) + \frac{(-\nu)(-\nu+1)}{2}s(x-2,y) +$$
$$+ \frac{(-\nu)(-\nu+1)(-\nu+2)}{6}s(x-3,y) + ...$$
$$\frac{\partial^\nu s(x,y)}{\partial y^\nu} = s(x,y) + (-\nu)s(x,y-1) + \frac{(-\nu)(-\nu+1)}{2}s(x,y-2) +$$
$$+ \frac{(-\nu)(-\nu+1)(-\nu+2)}{6}s(x,y-3) + ...$$

(2)

We can also define the fractional gradient [8]:

$$\vec{\nabla}^\nu = \frac{\partial^\nu}{\partial x^\nu}\vec{u}_x + \frac{\partial^\nu}{\partial y^\nu}\vec{u}_y = G_x^\nu \vec{u}_x + G_y^\nu \vec{u}_y$$

(3)

where $\vec{u}_x, \vec{u}_y$ are the unit vectors of the two space directions. Given the two components of the gradient, we easily evaluate the magnitude $G^\nu = ((G_x^\nu)^2 + (G_y^\nu)^2)^{1/2}$. In [8], authors prefer approximating the magnitude with $G^\nu = |G_x^\nu| + |G_y^\nu|$. In the case when the fractional order parameter is $\nu = 1$, we have the well-know gradient.

**3. Fractional edge detection**
There are many methods for the edge detection of an image. Most of these methods are based on computing a measure of edge strength, usually an expression containing the first-order derivatives, such as the gradient magnitude. Moreover, the local orientation of the edge can be estimated by means of the gradient direction. In the case of second order calculations, the methods use the Laplacian calculation. Recently, we have introduced dipole and quadrupole moments of the image maps, defined as in physics dipole and quadrupole moments of charge distribution are, and used them for image edge detection too [10-12].

The fractional differentiation was considered for edge detection in Ref.[13]. The authors demonstrated how the use of an edge detection based on non-integer differentiation improves the detection selectivity. They have also discussed the improved quality in term of robustness to noise. Unfortunately, Ref.[13] does not show any example of edge detection on images.

To have an idea of possible results that we can obtain with fractional order derivatives, let us try to apply Eq.2 and 3 to image maps. As in Ref.10-12, we prepare an output map as follow. The magnitude of the gradient $G^v(x,y,c)$ is evaluated on the function $s(x,y)$ given by the image map $b(x,y,c)$ for each colour tone $c$. Partial derivatives contain only the first four terms in Eq.2 and 3. For each colour, we find the maximum value $G^v_{Max}(c)$ on the image map. Let us define the output map as in the following:

$$b_G(x,y,c) = 255 \left( \frac{G^v(x,y,c)}{G^v_{Max}(c)} \right)^\alpha \qquad (4)$$

where $\alpha$ is a parameter suitable to adjust the image visibility. Fig.1 reports the results obtained with three different values of fractional parameter $v$, using the original image (1.a). The value of $\alpha$ is fixed at 0.4. Map (1.b), obtained for $v=1$, behaves as a usual edge detection. Maps (1.c) and (1.d) are obtained with fractional values, $v=0.7$ and $v=0.3$, respectively.

Form Fig.1, we can see that the original image remains clearly visible, with a strong enhancement of its edges. Therefore, we can try to use the fractional gradient on blurred images. Fig.2 displays how the fractional calculus works on the same original image (1.a) subjected to a Gaussian blurring (2.a). Map (2.b) is obtained for $v=1, \alpha=0.7$ and maps (2.c) and (2.d) are obtained with $v=0.5, \alpha=0.7$ and $v=0.5, \alpha=0.8$, respectively. These maps display a certain focusing effect.

**4. A tool for astronomical images**

As we have seen, fractional differentiation gives a quite different approach to edge detection. Let us remember that edge detection plays a fundamental role in texture segmentation: we can then imagine image segmentation based on fractional edge detection. In fact, such a method has been already proposed in Ref.[14]. Studies on fractional Fourier transformations are also possible, such as an image processing based on Green's function solutions of fractional diffusion equation [15]. This reference is quite interesting because it is proposing a method to remove from images the incoherent light scattering produced by a random medium. The author shows examples of quality enhancement of astronomical images.

The implementation of the approach proposed in Ref.15 is not immediate. We prefer the following more simple approach, which consists in testing the effect of the fractional gradient (3) on astronomical images. Fig.3.a shows Saturn observed with three different narrow-band filters (authors J. Näränen and R. Karjalainen, Nordic Optical Telescope Scientific Association). (3.b), (3.c) and (3.d) are the maps obtained from the magnitude of gradient $G^v$, for $v=0.7$, $v=0.6$ and $v=0.4$ respectively. For these three images, we set $\alpha=0.3$. In the output images, we can see five satellites. The increase of contrast and brightness of image (3.a) is not able to give a result with the same quality, in the detection of faint objects in the background field.

Another example of image processing with fractional gradient is given in Fig.4. The upper part of the figure shows Messier 45 star cluster, the Pleiades (author is Paul Milligan). The lower image shows the result of fractional differentiation with $\nu = 0.7$ and $\alpha = 0.4$. In this case, we performed a slight adjustment of brightness and contrast.

Here again, we have an enhanced visibility of feeble objects. As a result, we suggest that fractional differentiation could properly enter those image-processing tools devoted to the detection of faint objects in astronomical images. In spite of all the open questions concerning the application of fractional calculus to image processing [8], this research area surely deserve a more rich attention, as shown by the interesting performances of the fractional gradient.

ACKNOWLEDGEMENT
Many thanks to Paul Milligan, http://www.eyetotheuniverse.com/, for the kind permission of using the Pleiades image for fractional elaboration.

**Figure captions**

**Fig.1**. Image shows the result of fractional differential calculation on an image. Image (a) shows the original image. (b) is the magnitude of the gradient $G^1$. (c) and (d) are representing the magnitude of gradient $G^\nu$, for $\nu = 0.7$ and $\nu = 0.3$, respectively. Parameter $\alpha$ is equal to 0.4.

**Fig.2**. Image displays how the fractional calculus works on an image subjected to a Gaussian defocusing (a). Map (b) is obtained for $\nu = 1, \alpha = 0.7$ and maps (c) and (d) are obtained with $\nu = 0.5, \alpha = 0.7$ and $\nu = 0.5, \alpha = 0.8$, respectively.

**Fig.3**. Image (a) shows Saturn observed with three different narrow-band filters. The authors are J. Näränen and R. Karjalainen, Nordic Optical Telescope Scientific Association. (b), (c) and (d) are the maps obtained from the magnitude of gradient $G^\nu$, for $\nu = 0.7$, $\nu = 0.6$ and $\nu = 0.4$ respectively. For the three images, we have $\alpha = 0.3$. Note how the fractional calculation puts in evidence five satellites.

**Fig.4**. The upper part of the figure shows Messier 45 star cluster, the Pleiades (image obtained by Paul Milligan; location: Isle of Man, British Isles). The lower image shows the result of fractional differentiation with $\nu = 0.7$ and $\alpha = 0.4$ (with a slight adjustment of brightness and contrast).

High resolution figures at  http://staff.polito.it/amelia.sparavigna/fractional/

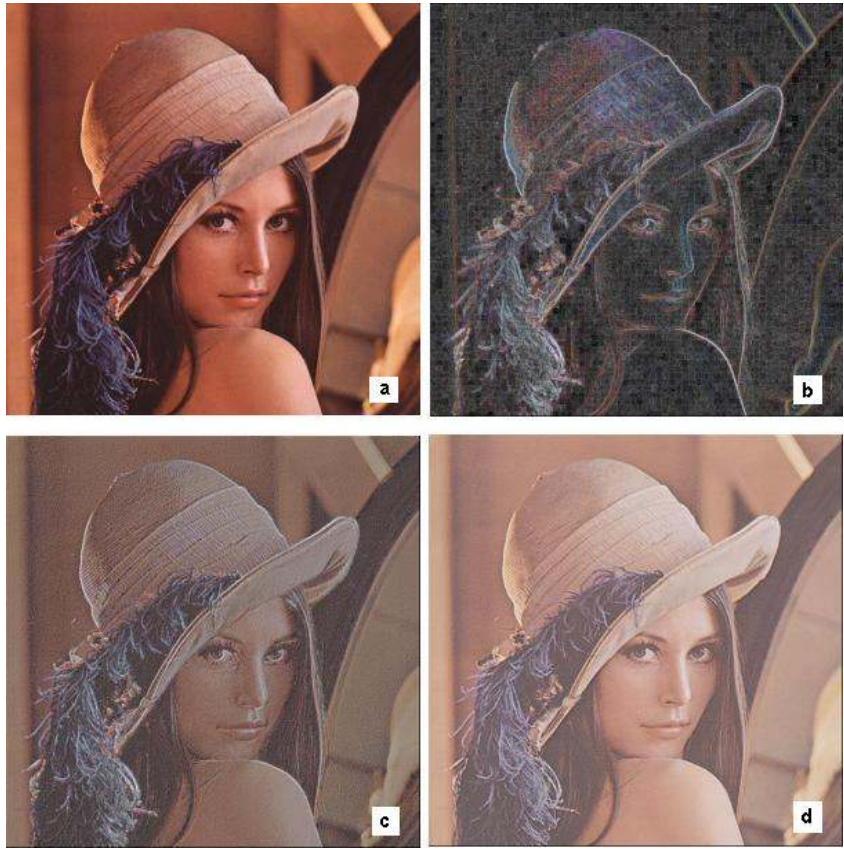

Fig.1

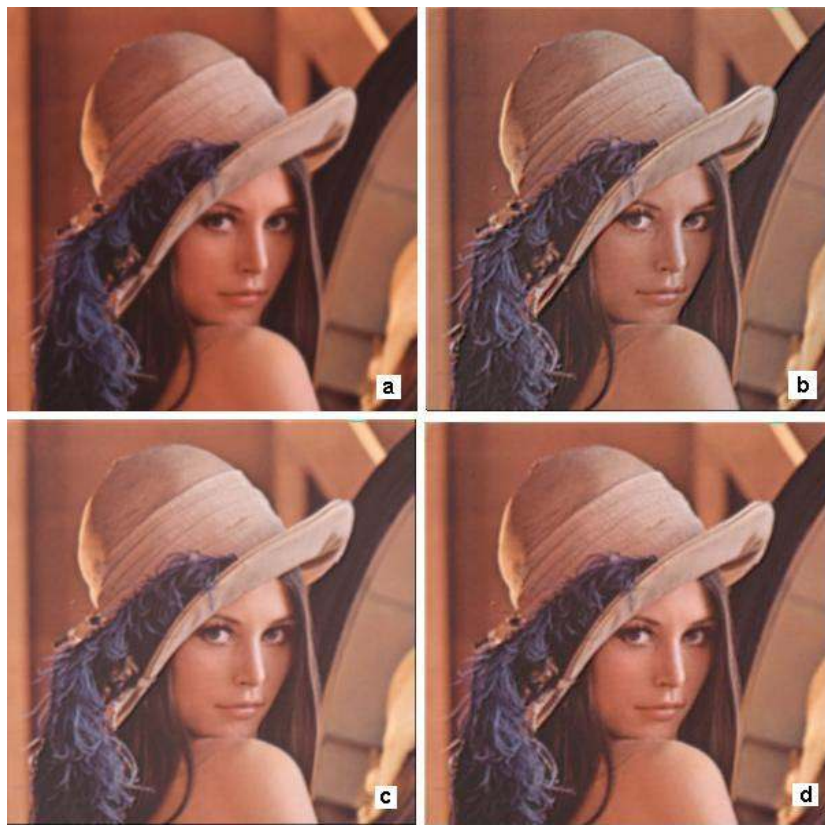

Fig.2

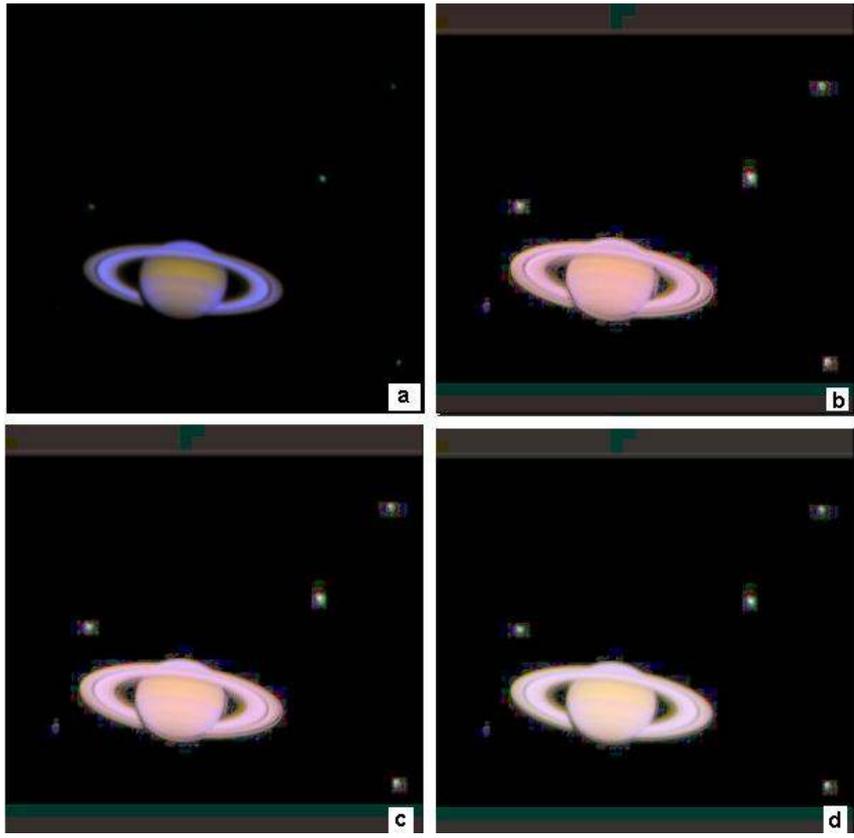

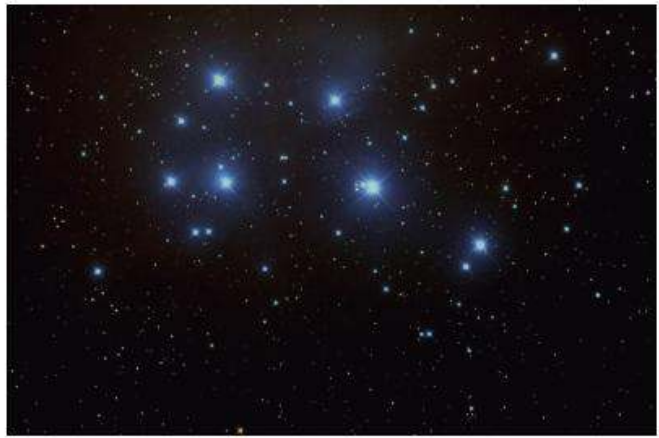

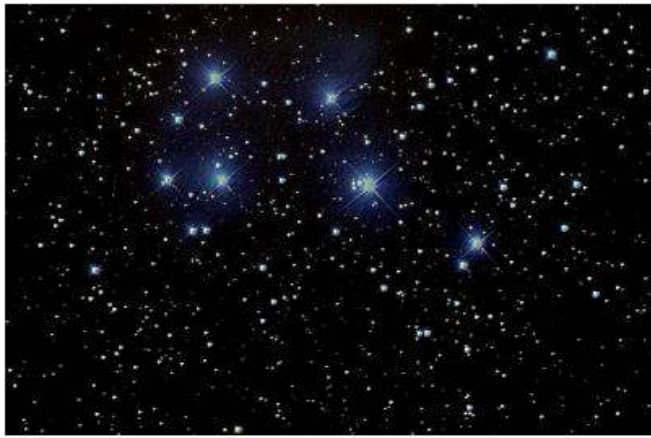

Fig.4